\documentclass[11pt,a4paper]{article}
\usepackage[hyperref]{acl2020}
\usepackage{times}
\usepackage{latexsym}
\pdfoutput=1 
\usepackage{graphicx}

\usepackage{microtype}
\usepackage{etoolbox}

\aclfinalcopy % Uncomment this line for the final submission
\title{A Brief History of Named Entity Recognition}

\author{Monica Munnangi \\
  University of Massachusetts, Amherst \\
  \texttt{mmunnangi@cs.umass.edu}  \\}

\date{}

\begin{document}

\maketitle

\begingroup\def\thefootnote{*}\footnotetext{ Survey performed in 2020.}

\begin{abstract}
A large amount of information in today’s world is now stored in knowledge bases.
Named Entity Recognition (NER) is a process of extracting, disambiguation, and linking an entity from raw text to insightful and structured knowledge bases. More concretely, it is identifying and classifying entities in the text
that are crucial for Information Extraction, Semantic Annotation, Question Answering, Ontology Population, and so on. The process of NER has evolved in the last three decades since it first appeared in 1996. In this survey, we study the evolution of techniques employed for NER and compare the results, starting from supervised to the developing unsupervised learning methods.
\end{abstract}

\section{Introduction}
\label{sec:intro}

Named Entity Recognition refers to the task of identifying the names of all people, organizations, and geographic locations in the given text, etc. Figure 1 shows an example task as shown in \cite{DBLP:journals/corr/abs-1812-09449}. It has now evolved to include Medical Codes, Time Expressions, quantities, and monetary value. The process of lifting Natural Language to knowledge Graph as covered in \citep{8999622} involves three steps :
\begin{enumerate}
    \item Named Entity Recognition (NER)
    \item Named Entity Disambiguation (NED)
    \item Named Entity Linking (NEL)
\end{enumerate}  
Most of the papers on Entity Extraction follow these steps mentioned above sequentially, but in the recent advances NER and NED have been done jointly, as they capture more information, this is termed as Named Entity Recognition and Disambiguation (NERD) as detailed in \citep{habib_van_keulen_2016} and \citep{Nguyen2016JNERDJN}. \\

\begin{figure}
\includegraphics[scale=0.65]{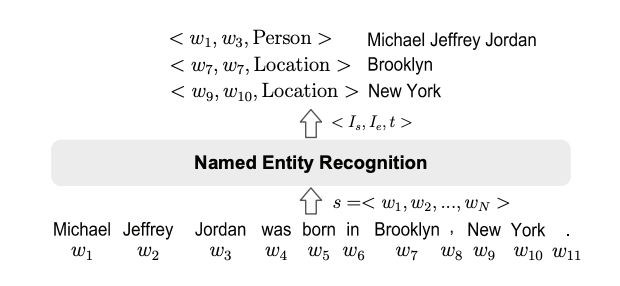}
\caption{Example Named Entity Recognition task \cite{DBLP:journals/corr/abs-1812-09449}}
\end{figure}

\textbf{Motivation for the survey :} In recent years, deep learning has attracted significant attention due to its success in various domains. DL-based NER systems with minimal feature engineering have given great results. Over the last decade, a decent number of studies have applied deep learning to NER which, advanced the state-of-the-art performance. We compare the choices of Deep Learning (DL) architectures to identify factors affecting NER performance as well as challenges that come with them.

Early NER systems were based on rule-based methods, lexicons, and ontologies. Machine Learning models have started around 2007, were based on feature-engineering. These models were domain-specific and did not generalize well when they were presented to new data.  Deep learning models and especially Recurrent Neural Networks (RNNs), are based on characters, sub-words, and word embeddings. Deep learning models proved to be more robust to new domains and new data. While these generalize well on new data, they need a lot of manually annotated data, which consumes a lot of time and resources which proves to be a bottleneck in a lot of cases.\\

Active learning, semi-supervised, and unsupervised learning techniques are methods that do not need human-annotated data but still give comparable results as deep learning models \cite{DBLP:journals/corr/ShenYLKA17}. We understand and compare the results on the CoNLL-2002 and CoNLL-2003 datasets, and 2010 i2b2/VA challenge data for the case study.\\

In the rest of the paper, We look into the most popular data-sets and tools to evaluate NLP tasks in section II and evaluation metrics in sectinon III. We discuss the background starting from Named Entity Recognition to building the initial Knowledge Graphs in Section IV. Section V constitutes the bulk of the document and reviews recent approaches to lifting these named entities in Natural Language texts to Knowledge Graphs. In Section VI, talk about unsupervised learning and discuss future directions. \\

%\begin{figure*}
%\includegraphics[width=\textwidth]{datasets.pn%g}
%\caption{Sample NER task}
%\label{fig:cclogo}
%\end{figure*}

\begin{figure*}
\includegraphics[width=\textwidth,height=7cm]{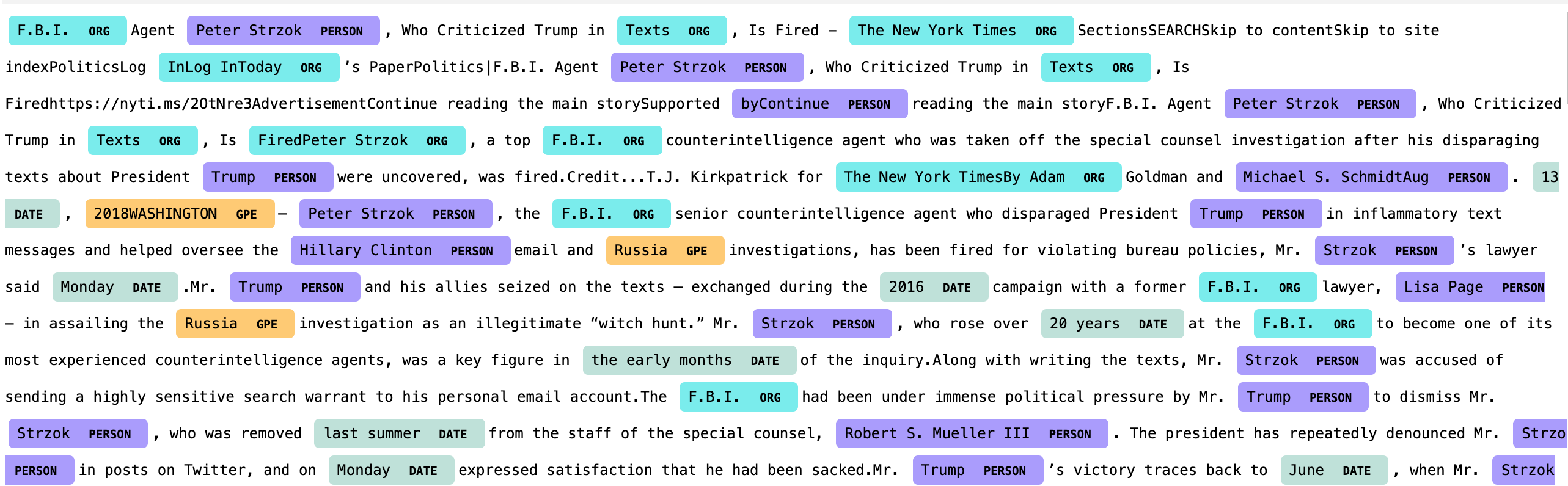}
\caption{Named Entity Recognition on New York Times dataset}
\label{fig:cclogo}
\end{figure*}

\section{Data-sets and Tools}

High quality data-sets are essential for both model evaluation and learning. Most of the them are news articles that were widely available on the internet and also Wikipedia articles. 
Twitter data became the next major data-set \cite{Owoputi2012PartofSpeechTF} to do parts-of-speech tagging but it had come with its own challenges of dealing with emoticons, sarcasm detection. 

\begin{enumerate}
   \item \href{https://catalog.ldc.upenn.edu/LDC2003T13}{MUC - 6} : 
   The corpus contains the 318 annotated Wall Street Journal articles, the scoring software and the corresponding documentation used in the MUC6 evaluation. Both the MUC 6 Additional News Text and the MUC 6 corpus are necessary in order to replicate the evaluation.
   \item \href{https://www.clips.uantwerpen.be/conll2003/ner/}{CoNLL03} :
   The task in CoNLL-2003 is the language-independent named entity recognition. It has four entities: persons, locations, organizations and names of miscellaneous entities that do not belong to the previous three groups. This data is used to develop a named entity recognition system.
   \item \href{https://catalog.ldc.upenn.edu/LDC2008T19}{NYT} :
    The text in this corpus is formatted in News Industry Text Format (NITF) developed by the International Press Telecommunications Council, an independent association of news agencies and publishers. The goals of NITF are to answer the essential questions inherent in news articles: Who, What, When, Where and Why. It includes over 1.8 million articles, Over 650,000 article summaries written by library scientists and over 1,500,000 articles manually tagged by library scientists with tags drawn from vocabulary of people, organizations, locations and topic descriptors.
   \item \href{https://catalog.ldc.upenn.edu/LDC2013T19}{OntoNotes} :
   Documents describing the annotation guidelines and the routines for deriving various views of the data from the database are included in the documentation directory of this release.
   \item \href{http://www.geniaproject.org/genia-corpus}{GENIA} :
   The GENIA corpus is the primary collection of biomedical literature compiled and annotated exclusively for the project. The corpus was created to support the development and evaluation of information extraction and text mining systems in molecular biology.
   \item \href{https://www.ncbi.nlm.nih.gov/CBBresearch/Dogan/DISEASE/}{NCBI - Disease} :
   The NCBI disease corpus is fully annotated at the mention and concept level to serve as a research resource for the biomedical natural language processing community. It has 793 PubMed abstracts with 6,892 disease mentions and a large number of unique disease concepts (790). 91\% of the mentions map to a single disease concept divided into training, developing and testing sets.  \cite{10.5555/2772763.2772800}.

\end{enumerate}

Training models from scratch takes a lot of time and computational resources. Therefore, there are many NER tools available online with pre-trained models namely, NeuroNER, StanfordCoreNLP, OSU Twitter NLP, Illinois, NLP, and NERsuite are offered by academia. spaCy, LingPipe, NLTK, OpenNLP, AllenNLP are from industry or open source projects. The following are most widely used tools and pretrianed models for Named Entity Recognition. 

\begin{enumerate}
    \item NeuroNER \cite{DBLP:journals/corr/DernoncourtLS17} : Named-entity recognition (NER) aims at classifying entities of mentioned in the text, such as location, organization and temporal expression.
    It leverages the state-of-the-art prediction capabilities of deep learning like Artificial Neural Network. It is cross-platform, open source, freely available.
    \item AllenNLP \cite{Gardner2017AllenNLP}
     : The named entity recognition model identifies named entities such as people, locations, organizations, etc. in the input text. The baseline in this model is based on ELMO \cite{DBLP:journals/corr/abs-1802-05365}. It uses a Gated Recurrent Unit (GRU) character encoder as well as a GRU phrase encoder, and it starts with pretrained GloVe vectors for its token embeddings and trained on the CoNLL-2003 NER dataset.
    \item StanfordCoreNLP \cite{manning-etal-2014-stanford}: Recognizes named entities such as person and company names, etc. in text. Principally, this annotator uses one or more machine learning sequence models to label entities, but it may also call specialist rule-based components, such as for labeling and interpreting times and dates. It is also known as CRFClassifier. The software provides a general implementation of linear chain Conditional Random Field (CRF) sequence models. That is, by training your own models on labeled data, you can actually use this code to build sequence models for NER or any other task. 
    \item spaCy : This Named Entity Recognition system features a sophisticated word embedding strategy using sub-word features and "Bloom" embeddings, a deep convolutional neural network with residual connections, and a novel transition-based approach to named entity parsing. The system is designed to give a good balance of efficiency, accuracy and adaptability. 
\end{enumerate}

\section{Evaluation Metrics} 
Metrics that define how good the result of an algorithm is and this is needed for any statistical judgment between two systems/algorithms and so we have added this section before discussing any NER models in order to equip the reader with the most popular evaluation metrics used in named entity recognition. \\

A confusion matrix is a technique for summarizing the performance of a classification algorithm but this alone can be misleading if there is an imbalanced class in the training data. It gives a better idea of how classification model is getting right and what types of errors it's encountering in the process. The following are the categories in a confusion matrix for a binary classification problem.  \\

\begin{enumerate}
    \item True Positive (TP) : entities that are recognized by NER and match the ground truth.
    \item False Positive (FP) : entities that are recognized by NER but do not match ground truth.
    \item True Negative (TN) : entities that predict the negative class correctly.
    \item False Negative (FN) : entities annotated in the ground truth that are not recognized by NER. 
    
\end{enumerate}
    
    $$ Recall = \frac{TP}{{TP + FN}} $$ 

    $$ Precision = \frac{TP}{{TP + FP}} $$ 
    
\begin{figure}
\centering
\includegraphics[scale=0.15]{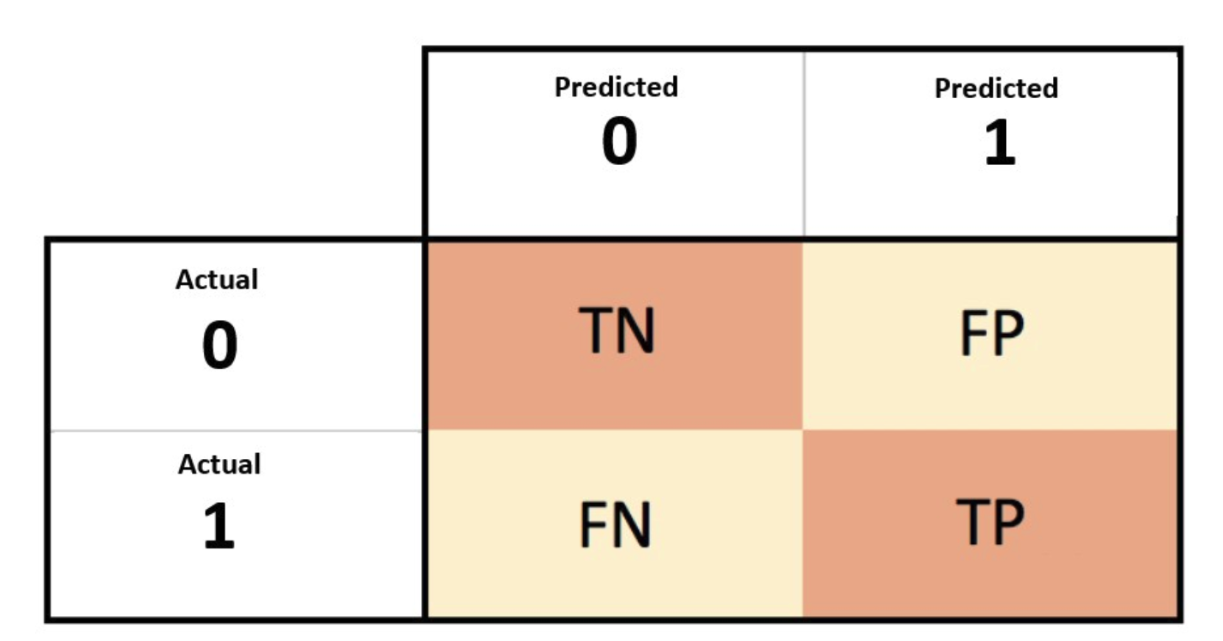}
\caption{Confusion Matrix}
\end{figure}

It is difficult to compare two models just with the precision and recall value combinations or vice versa. So to make them comparable, we use F-Score which is a measure of Recall and Precision at the same time. It uses Harmonic Mean in place of Arithmetic Mean by punishing the extreme values more.    

    $$ F1-measure = \frac{2 * Precision * Recall}{{Recall + Precision}} $$ 

Sensitivity measures how apt the model is to detecting events in the positive class. Specificity measures how exact the assignment to the positive class is. 

    $$ Sensitivity = \frac{TP}{{TP + FN}} $$ 

    $$ Specificity = \frac{TN}{{FP + TN}} $$ 

AUC - ROC curve is a another performance measurement for classification problem. ROC is a probability curve and AUC represents degree or measure of separability. It tells how much model is capable of distinguishing between classes. Higher the AUC, better the model is at predicting 0s as 0s and 1s as 1s.  \\

MUC-6 \cite{DBLP:journals/corr/abs-1812-09449}defined a relaxed-match evaluation: a correct type is credited if an entity is assigned its correct type regardless its boundaries as long as there is an overlap
with ground truth boundaries; a correct boundary is credited regardless an entity’s type assignment. Then ACE \cite{DBLP:conf/nlpcc/0101Y19} proposes a more complex evaluation procedure. It resolves a few issues like partial match and wrong type, and considers
subtypes of named entities. Complex evaluation methods are not intuitive and make error analysis difficult. Thus, complex evaluation methods are not widely used in recent studies. \\

CoNLL: Computational Natural Language Learning \cite{tjong-kim-sang-de-meulder-2003-introduction}, the Language-Independent Named Entity Recognition task introduced at CoNLL-2003 measures the performance of the systems in terms of precision, recall and f1-score, where:

``\textit{Precision} is the percentage of named entities found by the learning system that are correct. \textit{Recall} is the percentage of named entities present in the corpus that are found by the system. A named entity is correct only if it is an exact match of the corresponding entity in the data file." \\

BLEU score \cite{papineni-etal-2002-bleu}  is a method of automatic machine translation evaluation that is quick, inexpensive, and language-independent,
that correlates highly with human evaluation, and that has little marginal cost per run. \\

\begin{figure}[t]
\centering
\includegraphics[scale=0.32]{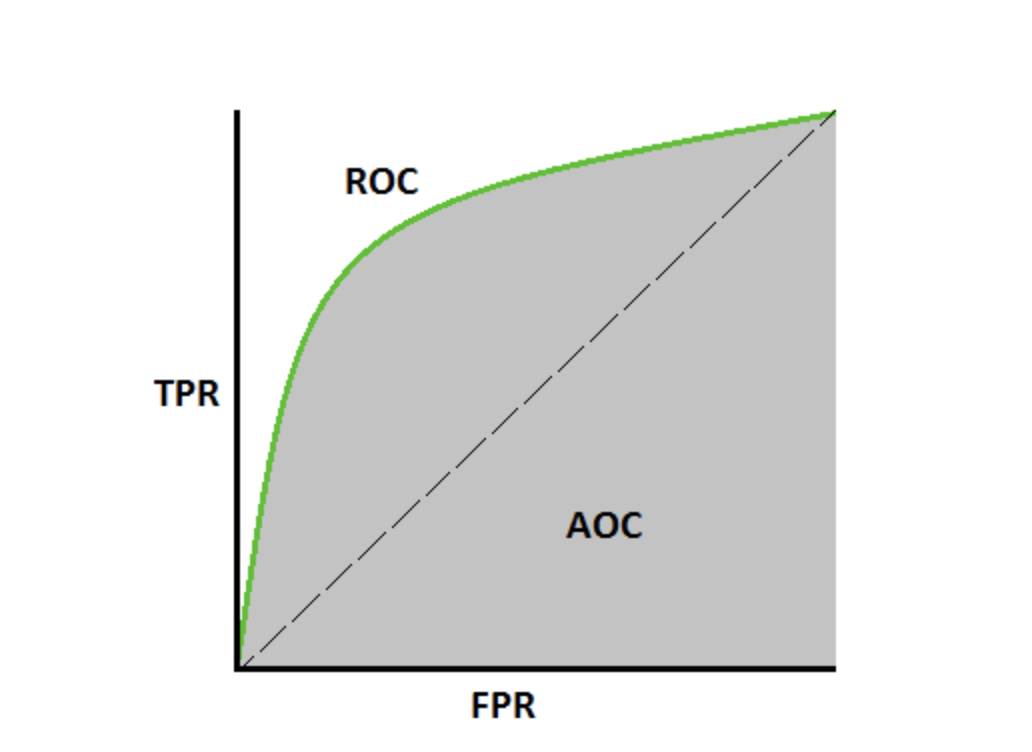}
\caption{An Example AUC - ROC Curve}
\end{figure}

These metrics are still relevant in machine learning and to evaluate models. NER had the following hard-tasks : 
\begin{enumerate}
    \item \textbf{Co-reference}: This occurs when two or more expressions in a text refer to the same person or thing
    \item \textbf{Word sense disambiguation}: A single work could have more than one meanings
    \item \textbf{Predicate-argument structure}: the system would have to create a tree interrelating the constituents of the sentence.

\end{enumerate}

\section{Background}

Named Entity Recognition first appeared in \citep{Grishman}, where the task involved identifying the names of all people, organizations, and geographic locations in the given text. The conference was an effort to promote and research in information extraction.  The first edition of MUC \cite{10.3115/1075671.1075684} took place in  1987 used ten narrative paragraphs from naval
messages as a training corpus and two others as test data and had no defined evaluation task or metrics. Researchers from six organizations ran their systems on the test data during the conference, then demonstrated and explained how the systems worked. In later editions, the performance metrics were developed, namely : precision and recall. This is a relaxed match evaluation; a correct type is assigned correctly even if it's partly correct. In simple terms, if the answer key has ${N_{key}}$ filled slots; and that a
system fills ${N_{correct}}$ slots correctly and ${N_{incorrect}}$
incorrectly. 

$$ Recall = \frac{{N_{correct}}}{{N_{key}}} $$ 

$$ Precision = \frac{{N_{correct}}}{{N_{correct}} + {N_{incorrect}}} $$ \\

\subsection{Rule-based methods}

Rule Based NLP methods rely on hand-crafted rules, designed by domain experts and lexical patterns. In the biomedical domain \cite{Hanisch2005ProMinerRP} uses a pre-processed synonym dictionary to identity protein mentions and potential genes in these texts. Other systems use hand-crafted semantic and syntactic rules to recognize entities. These are domain specific and are not an exhaustive list. Therefore, these systems result in high precision and low recall, and for the same reason, these cannot be used in cross domains.

\begin{figure}
\includegraphics[scale=0.7]{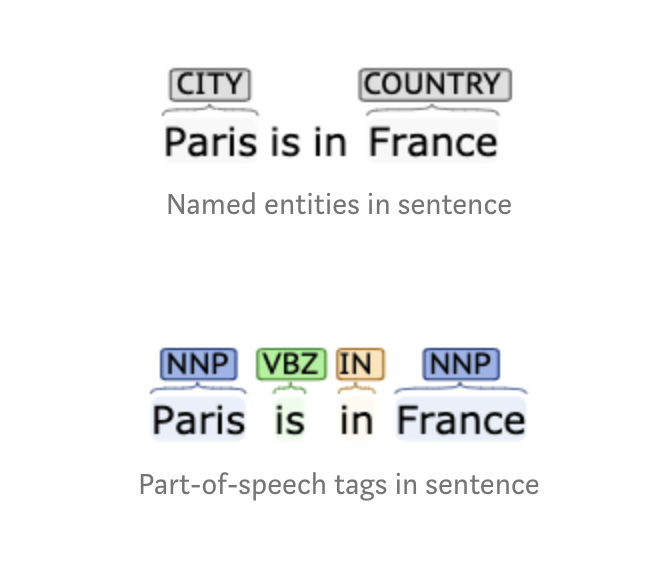}
\caption{Rule Based NER as described in \cite{chiticariu-etal-2010-domain}}
\end{figure}

Many instances of relations can be identified through hand-crafted patterns, looking for triples $ (X, \alpha, Y) $ where X are entities and $ \alpha $ are words in between. For the “Paris is in France” example, $ \alpha = $ ``is in”. This could be extracted with a regular expression.

Rule based methods have certain advantages, and 67\% industry vendors use them according to a research study by IBM \cite{chiticariu-etal-2013-rule}. Rule-based approaches are easy to comprehend, maintain and incorporate within various domains. \cite{chiticariu-etal-2010-domain} explains how a rule based methods called Named Entity Rule Language (NERL) could be used for different domains and say that NERL is specifically geared towards building and customizing complex NER annotations and makes it easy to understand a complex annotator that may comprise hundreds of rules.    \\

\subsection{Feature Extraction Methods}

\begin{figure*}
\includegraphics[width=\textwidth,height=6cm]{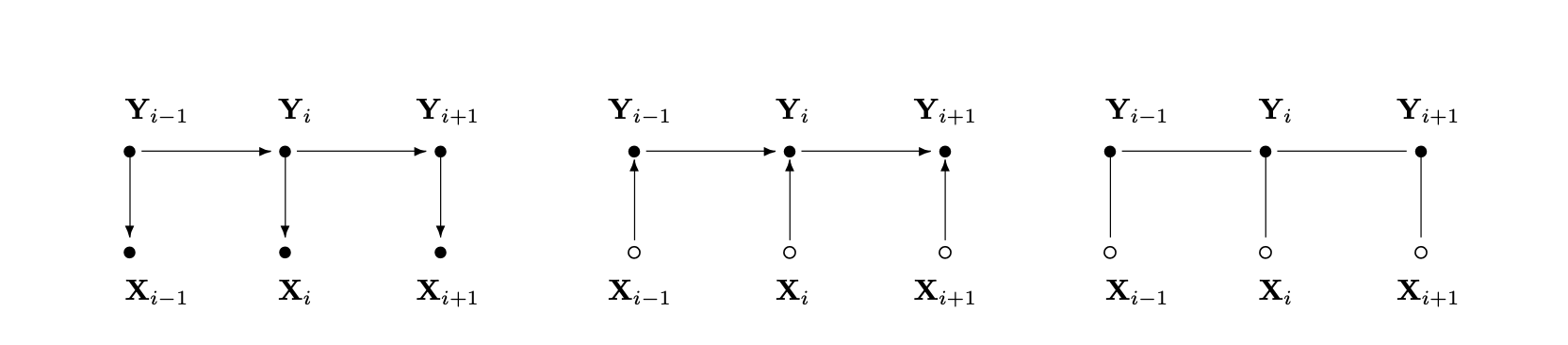}
\caption{Graphical structures of simple HMMs (left), MEMMs (center), and the chain-structured case of CRFs (right) for sequences.
An open circle indicates that the variable is not generated by the model as mentioned in \cite{Lafferty2001ConditionalRF}}
\label{fig:cclogo}
\end{figure*}

A step ahead are the supervised learning methods, which use feature engineering. Given the annotated data samples, features are carefully extracted and these are used to represent each data sample. Machine learning algorithms learn these patterns and use the methods on the unseen data to find the bucket they belong to, in case of classification. 
Most common machine learning models Decision trees \cite{10.1023/A:1022643204877}, Hidden Marcov Models (HMM) \cite{Rabiner86anintroduction}, Support Vector Machines and Conditional Random Fields (CRF) \cite{Lafferty2001ConditionalRF}. \\ 
Maximum entropy Markov models (MEMMs) \cite{10.5555/645529.658277} are conditional probabilistic sequence models that attain advantages such as looking at the past and future observations to predict the current words class, but they share a weakness called label bias problem, which is transition scores are conditional probabilities of possible next states give the current state and the observation sequence. 

As stated in \cite{Lafferty2001ConditionalRF}, HMMs and stochastic grammars are generative models, and had inherent disadvantages such ad assigning a joint probability to paired observation and label sequences. In particular, it is not practical to represent multiple interacting features or long-range
dependencies of the observations, since the inference problem for such models is intractable. CRF is a sequence modeling algorithm and this not only assumes that features are dependent on each other, but also considers future observations while learning a pattern combining the best of both HMM and MEMM. Conditional Random Fields is a class of discriminate models best suited to prediction tasks where contextual information or state of the neighbors affects the current prediction \cite{Lafferty2001ConditionalRF}. \\

\textit{Let $$ G = (V, E) $$ be a graph such that
$$Y = (Y_v)_{v \in V} $$ so that Y is indexed by the vertices
of G.}  \\

Then (X, Y) is a  conditional random field in
case, when conditioned on X, the random variables $Y_v$
obey the Markov property with respect to the graph:
$$ p(Y_v | X, Y_w, w \neq v) = p(Y_v | X, Y_w, w \sim v) $$
where $ w \sim v $ means that w and v are neighbors in G. \\

Log likelihood objective function for CRF : 

$$ O (\theta) = \sum_{i = 1}^{N} log\ P_{\theta} (y^{(i)} | x^{(i)}) $$
$$ \propto \sum_{x, y} p (x, y) log p_{\theta} (y | x) $$

CRF based NER has been applied in a multitude of domains, including tweets \cite{ritter-etal-2011-named}, for NER in foreign languages \cite{ekbal-etal-2008-named} and for chemical texts. \cite{10.1093/bioinformatics/btx761}. These feature selection machine learning models, cannot learn hidden features implicitly. These models fail especially, when the number of features increases exponentially. In the last decade, neural networks and deep learning became dominant and are achieving state-of-the-art results. We will discuss more about them in the next section. \\

\section{State-of-the-Art Systems}

Machine Learning includes a wide range of models, Support Vector Machine (SVM) \cite{Cortes95support-vectornetworks} was first introduced in 1995 and based on the idea of learning
a linear hyperplane that separates the one class from the other by a large margin. A large margin suggests that the distance between the hyperplane and the point from either instance is maximum. The points closest to hyperplane on either side are known as support vectors and was a well-researched method for NER. \cite{10.3115/1118853.1118873} was the first paper to use SVM for NER for every plausible transition of named entities and tested on CoNLL-2003.
– Isozaki and Kazawa (2002) learned four SVMs
for one entity type. It showed that SVM based
system was better than both HMM-based and
rule-based systems.

Deep learning, as defined in \cite{Goodfellow-et-al-2016}, a central problem in representation learning by introducing representations that are expressed in terms of other, more straightforward representations. Deep learning enables the computer to build complex concepts out of simpler concepts. \\

There are three core benefits of using deep learning for NER problems as they generate non-linear mappings from input to output : 
\begin{enumerate}
    \item Deep learning models learn complex features from data via non-linear activation functions. 
    \item The models save effort om designing NER features; these do not need considerable domain expertise to annotate the data that is fed. Deep learning models are effective in learning useful but hidden representations from raw data. 
    \item These models can be trained end-to-end by gradient descent, which works on the concept of back-propagation and chain rule methods. 
\end{enumerate}

Convolutional Neural Networks (CNN)  \cite{DBLP:journals/corr/abs-1103-0398} can be used in NER, called a sentence approach network where a word is tagged with the consideration of whole sentence. Each word in the current sentence is embedded in an N-dimensional vector. A CNN can then be used to produce local features around each word and then global features for the whole sentence. Bio-NER was proposed by \cite{doi:10.1093/bioinformatics/bty869} for biomedical named entity recognition. 

Supervised NER systems, including deep learning models, require large, annotated data for training. However, the process of data annotation is time-consuming and expensive \cite{DBLP:journals/corr/abs-1812-09449}. The annotations have to carefully done to avoid any mislabeling and avoid burnout of the annotator. It is tough for resource-poor languages and domains with little data available to perform well. For the same reason, the annotated for one domain or language cannot be used for another set of data or domain. Sometimes it might not even work on a different data-set within the same domain. \\

\cite{DBLP:journals/corr/abs-1903-07785} presents a new approach for pre-training a bi-directional transformer model that provides
significant performance gains across a variety of language understanding problems, the paper implements a biLSTM-CRF from  \cite{DBLP:journals/corr/abs-1802-05365} with two minor modifications: (1) instead of two layers of biLSTM, used only one, and (2) a linear projection layer was added between the token embedding and
biLSTM layer. \cite{panchendrarajan-amaresan-2018-bidirectional} describes the Bidirectional LSTM with CRF for the NER task. Instead of a single CRF CRF, they use BI-CRF to capture the dependencies between labels in both rights and left directions of a sequence.

\begin{figure}
\includegraphics[scale=0.6]{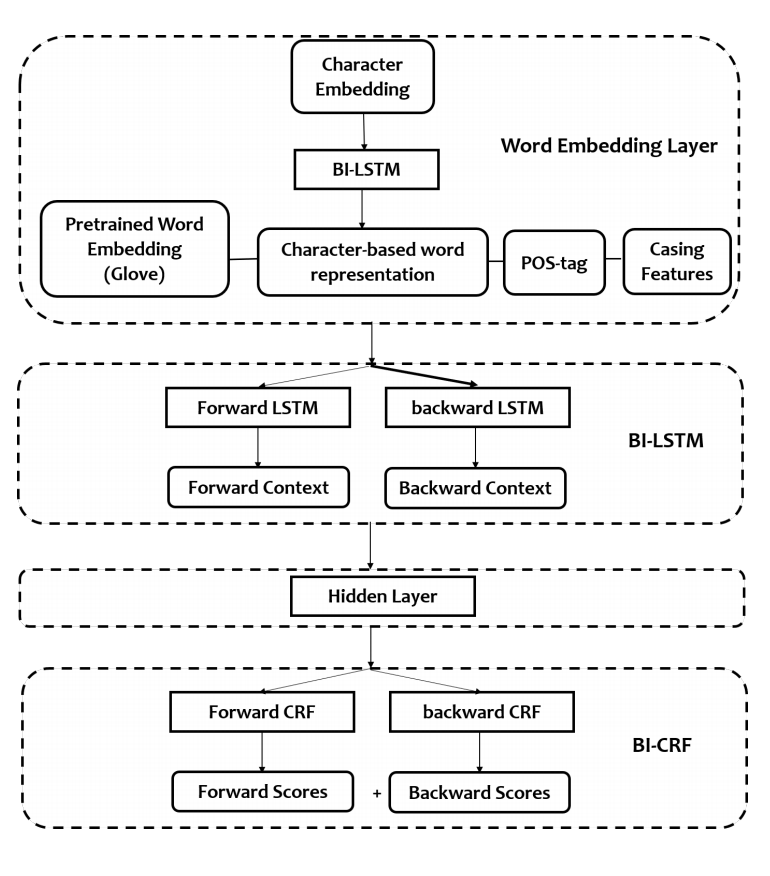}
\caption{BiLSTM - CRF architecture as mentioned in \cite{DBLP:journals/corr/abs-1903-07785} }
\end{figure}

% \begin{figure}
% \includegraphics[scale=0.4]{results.png}
% \caption{Average F1 scores of individual labels in \cite{DBLP:journals/corr/abs-1903-07785}}
% \end{figure}

\begin{enumerate}
    \item Character-based word representation is learned using a BI-LSTM, up table representing the character embedding is randomly initialized with the size of all possible characters. These concatenated with POS tag and casing features to obtain the final embedding representation of a word.
    \item Dropout is applied to the embedding layer, which is a regularization technique for reducing overfitting in neural networks by randomly dropping neurons in the network with a certain probability. This encourages the next layer (BI-LSTM) to utilize all four features of this layer to learn the contextual vector representation of a word in the next layer.
    \item This variation of the LSTM is referred to as bidirectional LSTM (BI-LSTM). Here, the input is given to forward and backward LSTMs to capture both the left and right context of the word. The final representation of a word is obtained by concatenating the left context.
    \item As the hidden layer on top of the BI-LSTM produces the score matrix P for a given sequence; the CRF layer learns only the transition probability of the output labels, $ A \in RK+2×K+2 $ K is the number of distinct labels, and +2 indicates one tag each for start and end marker. 
    \item During training, the negative sum of log probability of the correct sequences of both forward $ CRF (ln p(y | x)F ) $ and backward $ CRF (ln p(y | x)B ) $ is minimized. Different variations of combining the forward CRF and backward CRF results like maximum and average have experimented.
\end{enumerate}

\begin{figure}
\includegraphics[scale=0.2]{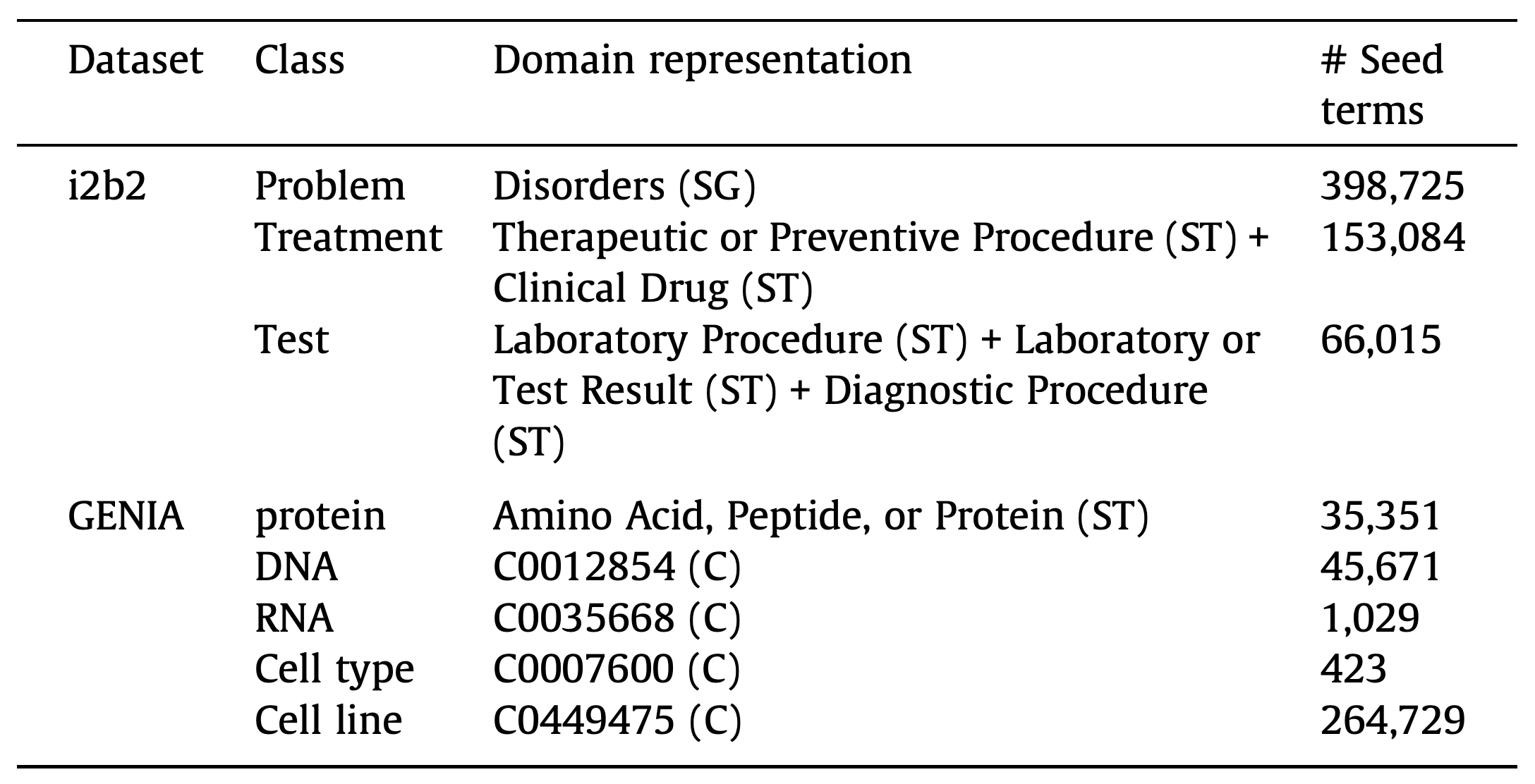}
\caption{Domain representations for entity classes in the i2b2 and GENIA corpora (ST:
semantic type; SG: semantic group; C: concept) \cite{journals/jbi/ZhangE13}}
\end{figure}

\section{Unsupervised and Beyond}
\label{sec:Unsupervised}
As discussed, the annotations are especially expensive to for NER since it requires multi-stage pipelines with sufficiently well-trained annotators. \cite{DBLP:journals/corr/ShenYLKA17} Active learning offers a solution for this problem, where it efficiently selects the set needed to label. As in supervised learning problem, where examples are drawn and labeled at random, in the active learning setting, the algorithm  choose which examples to  be labeled. Active learning aims to select a more informative set of examples instead of randomly drawing samples as in the supervised learning conditions. A central challenge in active learning is to determine what data is more informative than the rest and how the active learner can recognize this based on what it already knows. \\

Active learning offers a solution for this problem, where it efficiently selects the set needed to label. As in supervised learning problem, where examples are drawn and labeled at random, in the active learning setting, the algorithm  choose which examples to  be labeled. Active learning aims to select a more informative set of examples instead of randomly drawing samples as in the supervised learning conditions. \\

The first unsupervised algorithm was based on decision tree from \cite{yarowsky-1995-unsupervised} and the paper talks about DL-CoTrain, the input to the unsupervised algorithm is an initial "seed" set of rules. The method uses a "soft" measure of the agreement between two classifiers
as an objective function; we described an algorithm which directly optimizes this function. \\

\section{Future Directions}

Though NER is considered only a preprocessing method; however, with the advancement in Deep Learning models, it has become an active area of research, with improvements in semi-supervised and unsupervised learning. \\

With manual annotations becoming a bottleneck in a multitude of applications, as discussed in section~\ref{sec:Unsupervised}, there is a great scope of implementing semi-supervised and active learning methods. 
Domain adaptation \cite{10.1007/978-3-319-49409-8_35} is a field associated with deep learning, when we aim at learning from a source data distribution a well-performing model on a different but related target data distribution. The future direction could be making systems that can be used in different domains where data acquisition is laborious, and manual annotations are expensive. \\

In linguistics, a model trained on one data-set may not generalize and work well on other domain texts due to the differences in linguistics as well as the differences in annotations. However, there are some studies of applying deep transfer learning to NER \cite{ner-sekine2007}; this domain has not been fully explored. Prospective future work could be (a) developing a  robust system, which can generalize across different domains; (b) exploring zero-shot, one-shot, and few-shot learning in NER tasks; (c) providing solutions to address domain mismatch, and label mismatch in cross-domain settings. \cite{DBLP:journals/corr/abs-1812-09449}

\section*{Acknowledgments}

I thank Prof. Brendon O'Connor for providing the opportunity and TAs Varun and Shufan for their guidance and Shaurya for his valuable feedback. 

\bibliography{anthology,acl2020}
\bibliographystyle{acl_natbib}

\end{document}